\documentclass[3p,times,procedia]{elsarticle}
\usepackage{ecrc}
\usepackage[bookmarks=false]{hyperref}
    \hypersetup{colorlinks,
      linkcolor=blue,
      citecolor=blue,
      urlcolor=blue}
\usepackage{float}      
\usepackage{amsmath}
\usepackage{makecell}
\usepackage{multirow}

\volume{00}

\firstpage{1}

\journalname{Procedia Computer Science}

\runauth{H. Li et al.}

\jid{procs}

\usepackage{amssymb}

\usepackage[figuresright]{rotating}

\usepackage{subfigure}

\begin{document}
\begin{frontmatter}

\dochead{Proceedings of International Conference on Biomimetic, Intelligence and Robots}

\title{Semi-supervised Learning for Segmentation of Bleeding Regions in Video Capsule Endoscopy}

\author[a]{Hechen Li} 
\author[b,a]{Yanan Wu}
\author[a]{Long Bai}
\author[a]{An Wang}
\author[c]{Tong Chen}
\author[a,d,e,f]{Hongliang Ren\corref{cor1}}

\address[a]{Department of Electronic Engineering, The Chinese University of Hong Kong,
    Hong Kong, China}
\address[b]{College of Medicine and Biological Information Engineering, Northeastern University, Shenyang, China}
\address[c]{School of Electrical and Information Engineering, The University of Sydney, Sydney, Australia}
\address[d]{Shun Hing Institute of Advanced Engineering, The Chinese University of Hong Kong, Hong Kong, China}
\address[e]{Department of Biomedical Engineering, National University of Singapore, Singapore}
\address[f]{Shenzhen Research Institute, The Chinese University of Hong Kong, Shenzhen, China}
\begin{abstract}
In the realm of modern diagnostic technology, video capsule endoscopy (VCE) is a standout for its high efficacy and non-invasive nature in diagnosing various gastrointestinal (GI) conditions, including obscure bleeding. Importantly, for the successful diagnosis and treatment of these conditions, accurate recognition of bleeding regions in VCE images is crucial. While deep learning-based methods have emerged as powerful tools for the automated analysis of VCE images, they often demand large training datasets with comprehensive annotations. Acquiring these labeled datasets tends to be time-consuming, costly, and requires significant domain expertise. To mitigate this issue, we have embraced a semi-supervised learning (SSL) approach for the bleeding regions segmentation within VCE. By adopting the `Mean Teacher' method, we construct a student U-Net equipped with an scSE attention block, alongside a teacher model of the same architecture. These models' parameters are alternately updated throughout the training process. We use the Kvasir-Capsule dataset for our experiments, which encompasses various GI bleeding conditions. Notably, we develop the segmentation annotations for this dataset ourselves. The findings from our experiments endorse the efficacy of the SSL-based segmentation strategy, demonstrating its capacity to reduce reliance on large volumes of annotations for model training, without compromising on the accuracy of identification.
\end{abstract}

\cortext[cor1]{Corresponding author}
\end{frontmatter}

\email{hlren@ee.cuhk.edu.hk}



\section{Introduction}
\label{main}


Video capsule endoscopy (VCE) is a non-invasive diagnostic technique \cite{iddan2000wireless,zhang2022deep} that involves the patient swallowing a small capsule equipped with a miniature camera, recording videos of the gastrointestinal (GI) tract. VCE has become a valuable diagnostic tool for various GI conditions including GI bleeding, tumors, and Crohn's disease \cite{bai2023llcaps}. Accurate bleeding segmentation in VCE images is critical for diagnosing various GI disorders. However, this task is challenging because of several factors, including the variability in the appearance of bleeding (e.g., color, size, and shape), the existence of artifacts (e.g., debris or bubbles), and the influence of the capsule's motion on image quality \cite{soffer2020deep}. The lack of standardized criteria for bleeding identification contributes to inter- and intra-observer variability, potentially leading to inconsistent diagnoses and treatment recommendations \cite{postgate2009computer}.

To address these challenges, researchers have been developing advanced algorithms for automated bleeding segmentation in VCE images~\cite{aoki2020automatic,bai2022transformer}. Convolutional Neural Networks (CNNs), a subset of DL methodologies, have surfaced as a promising instrument for the interpretation of medical images~\cite{wu2021vision,wu2022two,wu2023deep,zhao2022cot,che2023image}. Several studies have reported accurate results and improved processing time when using CNNs for these tasks \cite{aoki2020automatic,jia2017gastrointestinal}. For instance, Aoki et al. \cite{aoki2020automatic} recognized blood content in VCE images automatically using a deep CNN. The study demonstrated improved performance compared to conventional image processing methods. Similarly, Jia et al. \cite{jia2017gastrointestinal} introduced an approach combining the CNN and handcrafted features for detecting GI bleeding in VCE, and it outperformed the existing CNN-based approaches in terms of specificity and sensitivity. However, CNN-based approaches often require large training datasets with extensive annotations and may suffer from limited generalizability across different patient populations and clinical settings \cite{iakovidis2015software}.

Semi-supervised learning (SSL) has gained recognition as a potent paradigm, leveraging both annotated and unannotated data in medical computer vision (CV) \cite{chebli2018semi}. The SSL approach is particularly valuable in medical imaging, where acquiring labeled data is always laborious, costly, and requires domain expertise \cite{litjens2017survey,bai2023surgical}. In recent years, SSL has demonstrated encouraging outcomes in diverse CV applications, for instance, tumor segmentation, disease classification, and lesion detection. The core principle of SSL lies in exploiting a vast amount of unlabeled data to enhance model generalization. By utilizing both annotated and unannotated data, SSL can capture the underlying patterns in the data more effectively, thereby leading to better representation learning \cite{oliver2018realistic}. Some popular SSL methods in medical CV include self-training \cite{xie2020self}, co-training \cite{peng2020deep}, multi-view learning \cite{yan2021deep}, and consistency regularization \cite{mustafa2020transformation}. These techniques serve as an effective countermeasure to the shortage of labeled data, fostering more dependable medical image analysis. The primary deliverables of this research include the following:
\begin{itemize}
    \item[--] We put forth a semi-supervised learning model for delineating bleeding regions in VCE.
    \item[--] We furnished the Kvasir-Capsule dataset with our own segmentation annotations.
     \item[--] Through thorough experimentation, we evidenced that our SSL model can significantly lessen the reliance on extensive, labeled datasets in healthcare settings.
\end{itemize}

\section{Methodology}

\subsection{Mean Teacher Method in Bleeding Segmentation}

The `Mean Teacher' \cite{tarvainen2017mean} is a semi-supervised learning approach for improving model generalization, in scenarios where labeled data is limited. Initially, a student and a teacher model, are created with the identical structure and initial weights. For the labeled data, a supervised loss is computed. Regarding the unannotated data, both the student and teacher networks are utilized to generate segmentation predictions. For example, we represent these two outputs for the same input $X_{unlabeled}$, plus different noise levels $\eta$ and $\eta^{\prime}$, as $f(X_{unlabeled}, \eta, \theta)$ and $f(X_{unlabeled}, \eta^{\prime}, \theta^{\prime})$, where $ \theta$ and $\theta^{\prime}$ represent the corresponding weights of the student and teacher models. The student aims to produce results similar to the teacher one for the unannotated data by computing an MSE consistency loss $L_{mse}$ between these two outputs. The student model incrementally picks up information from the teacher one via back-propagation through the minimization of the weighted loss of supervised loss $L_s$ and consistency loss $L_c$, as illustrated in Equation \ref{eq_semiloss}.

\begin{equation} \label{eq_semiloss}
\begin{aligned}
    L_{total} &= w_1 \, L_{s} + w_2\, L_{c} \\
    &= w_1 \, (L_{ce} + L_{dice}) + w_2\, L_{mse}
\end{aligned}
\end{equation}
Then, the teacher is updated utilizing the exponential moving average (EMA) of the parameters of the student, as shown in Equation \ref{EMA},
\begin{equation} \label{EMA}
    \theta^{\prime}_{t} = \beta\theta^{\prime}_{t-1}+(1-\beta)\theta_t
\end{equation}
where $\beta$ indicates a tunable EMA decay. The weights of these two models will be updated alternately until the model converges. 

\subsection{Modified U-Net with scSE Attention Block}

The Concurrent Spatial and Channel Squeeze \& Excitation (scSE) attention block \cite{roy2018concurrent} strengthens CNN feature representation by recalibrating spatial and channel-wise data. Comprising two parallel paths, the scSE attention block unifies the results of a CNN layer's output recalibrated through sSE and cSE branches.

The sSE path concentrates on spatial recalibration, employing a $1\times1$  convolution component and a subsequent sigmoid layer to generate a spatial attention map. This map, when multiplied element-wise with the inputs, emphasizes crucial spatial regions. On the other hand, the cSE branch targets channel-wise recalibration. It initially applies global average pooling to input feature maps to produce a channel-wise descriptor. This descriptor passes through two fully-connected components with a RELU function in the middle and a sigmoid function to yield channel-wise weights, which accentuate the most informative channels upon multiplication with the inputs.

The combination of both branches' outputs is achieved via an element-wise sum, leading to recalibrated feature maps incorporating both channel-wise and spatial attention. These recalibrated maps are then inputted into succeeding network layers.

\begin{figure*}[h]
\centering
\includegraphics[width=0.95\textwidth]{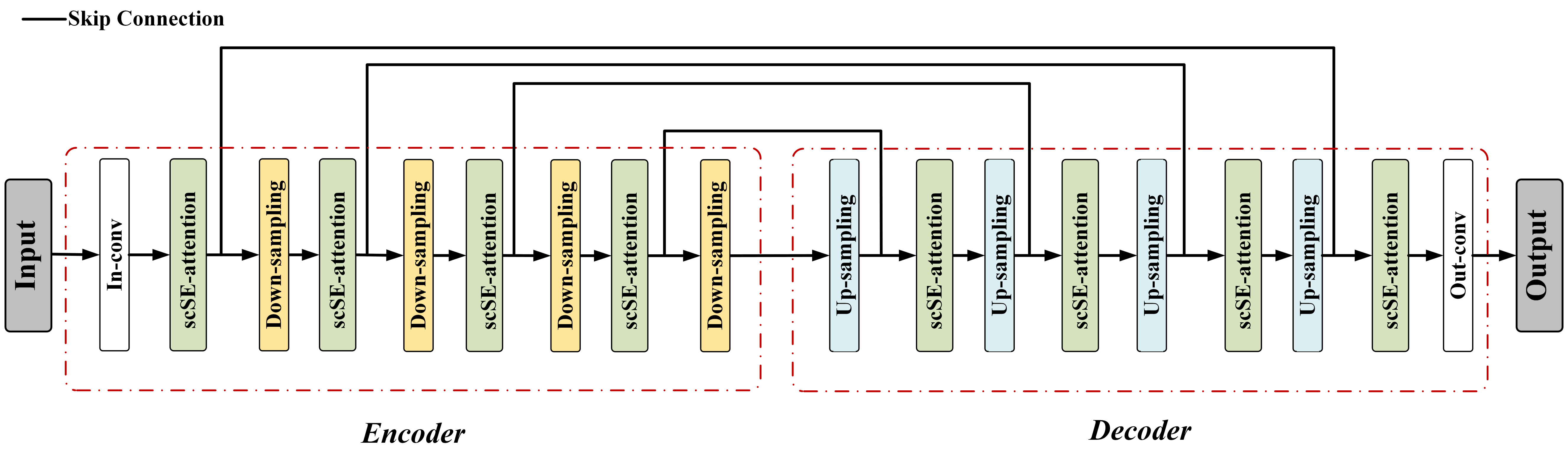}

\caption{\label{img_semi_net}The network structure of modified U-Net}
\label{architecture}
\end{figure*}

In the case of the student and teacher networks, as illustrated in Figure \ref{architecture}, we employ the modified U-Net \cite{ronneberger2015u} architecture enriched with the scSE attention mechanism. U-Net comprises an encoder and decoder architecture with bypass connections, and we integrate scSE attention modules post each down-sampling block in the encoder and following each up-sampling block in the decoder of the U-Net.

\section{Experiments}

\subsection{Dataset}

\begin{figure*}[htbp]
	\centering
	\subfigure[Detection annotation] {\includegraphics[width=.22\textwidth]{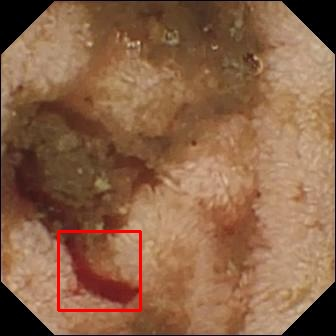}}\hspace{6mm}
	\subfigure[Segementation annotation] {\includegraphics[width=.22\textwidth]{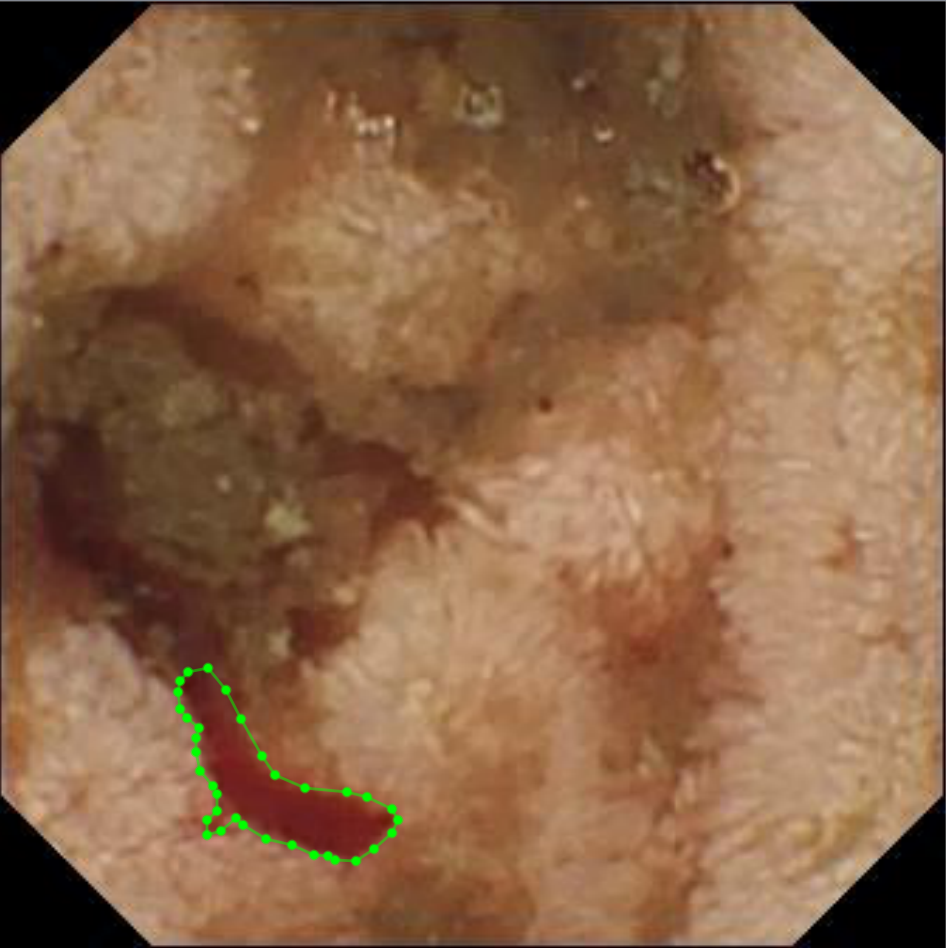}}\hspace{6mm}
	\subfigure[Binary mask image] {\includegraphics[width=.22\textwidth]{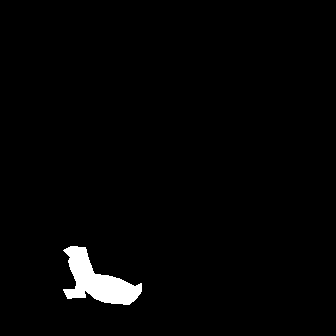}}
	\caption{Data annotation}
	\label{fig_2}
\end{figure*}

The dataset utilized in this paper is the Kvasir-Capsule dataset \cite{smedsrud2021kvasir}, providing VCE images collected during medical examinations at a hospital in Norway. The team has labeled and medically validated 47,238 frames from 117 videos, identifying findings in 14 different disease categories with bounding boxes. Among these categories, we selected the `Blood-fresh' category to meet our red lesions recognition task in VCE. The dataset comprises a total of 446 RGB images, each with a resolution of $336 \times 336$, sampled from seven video sequences. A ground truth bounding box, which indicates the detection, is provided for each frame. However, segmentation annotations are not available. Therefore, we created the segmentation annotation based primarily on the given detection annotation, as shown in Figure \ref{fig_2} (a), and sought guidance from a specialist when it was difficult to determine the presence of blood. As shown in Figure \ref{fig_2} (b), we used a series of points to meticulously outline the bleeding region using the LabelMe tool \cite{torralba2010labelme}. We generated a binary mask image for each image by utilizing the boundary coordinate information. The binary mask only contains pixel values of 0 and 1, with pixel value 0 indicating the background without blood and 1 denoting the presence of blood. The binary mask image is visualized in Figure \ref{fig_2} (c). We split the dataset according to the video sequences from different patients. The data from five patients were selected as our training set (389 images), while the data from the remaining two patients were chosen as our validation set (57 images).

\subsection{Implementation Details}

 All models are implemented using the PyTorch framework, with training conducted on two NVIDIA $\text{RTX}^\text{TM}$ 3090 GPUs over a total of 3000 iterations. The batch size is set as 16, divided equally between annotated and unannotated samples. To enhance model generalization in light of our limited data pool, data augmentation techniques including random flipping and rotation are applied. We utilize the SGD optimizer with an original learning rate of 0.01, adjusting it as per Equation \ref{lr},
\begin{equation} \label{lr}
    lr = lr_{base} \times (\frac{1 - c}{t}) ^ {0.9}
\end{equation}
where c indicates the present iteration number and t gives the total iterations of training.
During the ramp-up phase, we employ an EMA decay of $\beta$ = 0.99 in Equation \ref{EMA}, and for the remainder of the training,  $\beta $ = 0.999 is used. Recall the Equation \ref{eq_semiloss}, the weight $w_1$ for supervised loss is set to 0.5 and weight for consistency loss $w_2$ is ramped up from 0 to 1 during the first $L$ epochs using  Equation \ref{ramp},
\begin{equation} \label{ramp}
    w_2 = exp\, (-5\, (1-\frac{E}{L}) ^ {2})  
\end{equation}
where $E$ denotes the present training epoch, while $L$ can be defined as the ramp-up length, set as 50 in our experiment. When $ E > L$, $w_2$ is equal to 1. This ramp-up strategy ensures that the impact of consistency loss is limited during the initial training stages, as the teacher model may not produce accurate targets at an early age. Finally, we use the teacher model to perform the segmentation prediction.

\subsection{Results and Discussion}

\begin{figure*}[h]
	\centering
    \includegraphics[width=0.95\textwidth]{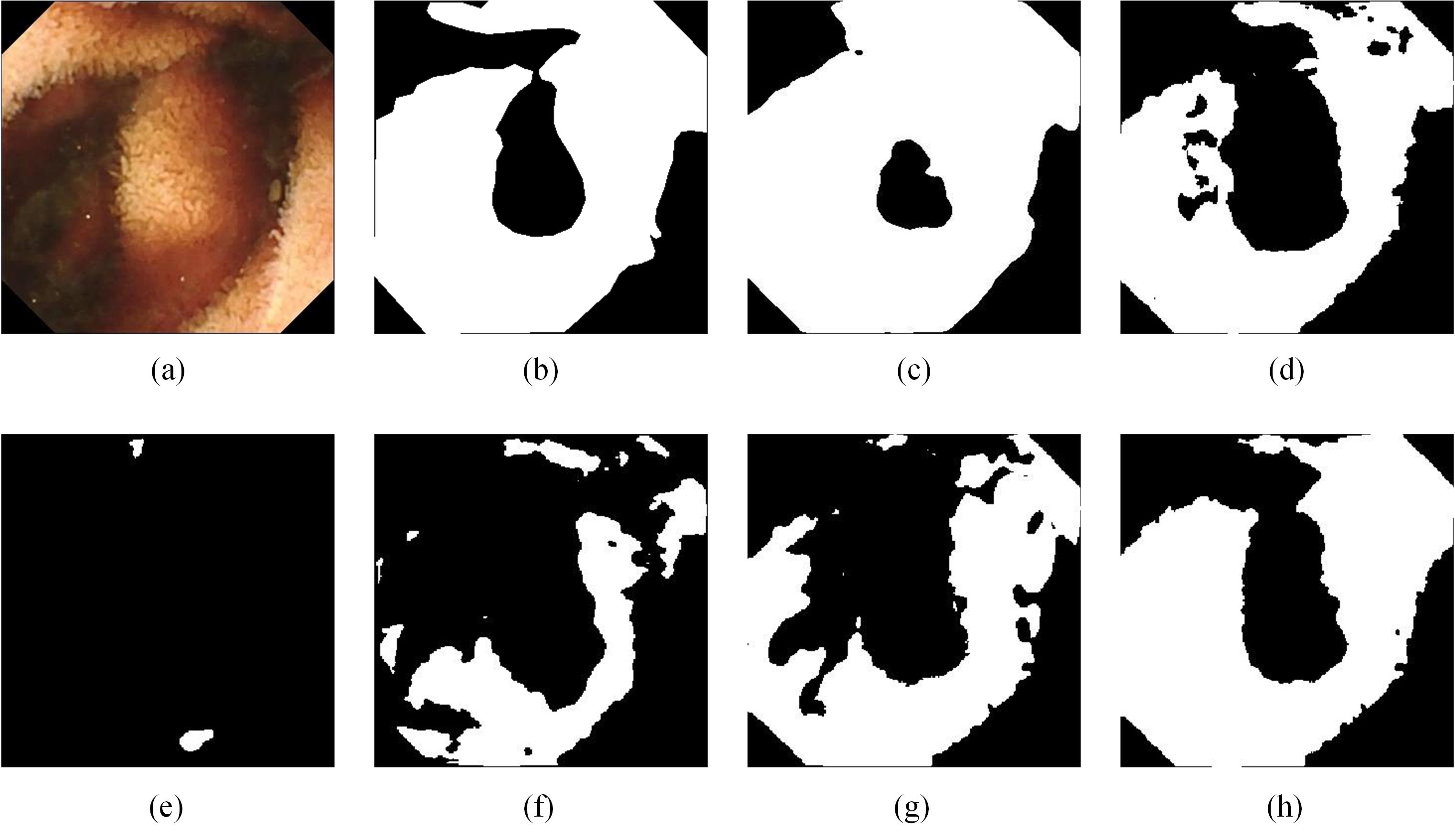}
	\caption{The influence of the number of given labels and whether the semi-supervised method is used on the segmentation results is compared. (a) Input Frame (b) Ground Truth (c) All labels with attention (d) All labels without attention (e) 50 labels-fully-supervised (f) 50 labels-semi-supervised (g) 150 labels-fully-supervised (h) 150 labels-semi-supervised}
	
	\label{vis_seg_comp}
\end{figure*}

The segmentation performance of all different models can be assessed using five common evaluation metrics including Dice score, mIoU, Sensitivity, Precision, and Hausdorff Distance (HD) respectively.
Firstly, we compare our scSE attention U-Net with three classical segmentation networks including E-Net \cite{paszke2016enet}, SegNet \cite{badrinarayanan2017segnet} and LinkNet \cite{chaurasia2017linknet} under all labels. The influence of attention blocks inserted in U-Net \cite{ronneberger2015u} is also assessed. Table \ref{tab_models} summarizes the outcomes of our comparative trials. Table \ref{tab_mode} conducts the experiments to investigate the effectiveness of our SSL strategy. We use different numbers of labels including 50, 100, 150 and all labels to evaluate our SSL model respectively. For each case, we train our model using fully-supervised and semi-supervised learning modes. Some segmentation results are visualized in Figure \ref{vis_seg_comp}. 

\begin{table*}[htbp]
\centering
\caption{\label{tab_models} Performance of different segmentation models using all labels: we compare our scSE attention U-Net with other segmentation networks under all labels.} \quad

\begin{tabular}{c c c c c c c}
\hline
Model & Dice & mIoU & Sensitivity & Precision & HD\\ \hline

E-Net \cite{paszke2016enet} & 0.6648 & 0.5692  & 0.6534  & 0.7315 & 43.0813 \\
U-Net \cite{ronneberger2015u} & 0.7590 & 0.6305  & 0.7727  & 0.8365 & 30.6598 \\ 
SegNet \cite{badrinarayanan2017segnet} & 0.7639 & 0.6472  & 0.7680  & 0.8180 & 35.0391 \\ 
LinkNet \cite{chaurasia2017linknet} & 0.7764 & 0.6544  & 0.7381  & \textbf{0.8794} & \textbf{26.5051} \\ 
Ours & \textbf{0.7845} & \textbf{0.6821}  & \textbf{0.8711}  & 0.7805 & 33.1743 \\\hline


\end{tabular}
\end{table*}

\begin{table*}[htbp]
\centering
\caption{\label{tab_mode} Performance of our proposed model using different training modes under different labels: we use different numbers of labels including 50, 100, 150 and all labels to evaluate our SSL model respectively.} \quad

\begin{tabular}{p{1.3cm}<{\centering} c c c c c p{1.5cm}<{\centering}}
\hline
Labels & Mode & Dice & mIoU & Sensitivity & Precision & HD\\ \hline

all & fully & 0.7845 & 0.6821  & 0.8711  & 0.7805 & 33.1743 \\ \hline
\multirow{2}*{50} & fully  &0.3736 &0.2858 &0.3538 &0.9011 & 92.4451  \\
		~ &  semi & 0.5536 & 0.4109 & 0.4352 & 0.9490 & 37.2052  \\ \hline
\multirow{2}*{100}  & fully & 0.6587 & 0.5093 & 0.5643 & 0.8989 & 35.8527 \\
~ & semi & 0.7086 & 0.5824 & 0.6174 & 0.9130 & 25.3400 \\ \hline
\multirow{2}*{150} & fully & 0.7215 & 0.5764 & 0.6567 & 0.8842 & 30.0753 \\
~ & semi & 0.7805 & 0.6639 & 0.8044 & 0.8158 & 30.3231 \\ \hline

\end{tabular}
\end{table*}

According to Table \ref{tab_models}, our model outperforms the three mentioned architectures under full-label training. Moreover, by comparing our model with the basic U-Net architecture, the role of scSE attention blocks is verified when the dice score increases from 0.759 to 0.785. 

As shown in Table \ref{tab_mode}, as more annotations are provided, the segmentation results improve quickly. The best result is obtained when using all labels for fully-supervised learning. After comparing the results under the same number of labels but using different training strategies, we can easily observe that the semi-supervised model surpasses its corresponding fully-supervised one obviously. Therefore, this experiment demonstrates the effectiveness of `Mean Teacher' method that we only use less than half of the labels to achieve nearly the same results as fully-supervised using all the labels. 

However, the overall segmentation performance may be relatively unsatisfactory. The potential reasons are analyzed as follows. Firstly, annotating the exact blood boundary in the tissue may be a difficult and subjective task. This can lead to inconsistent annotations which affect the accuracy of segmentation outcomes to some extent. Moreover, the dataset we used is quite small and has relatively less variation throughout each video sequence, and also the training and validation sets are quite different since the data varies widely from patient to patient.

\section{Conclusion}

Video capsule endoscopy has emerged as a beneficial tool for detecting and managing a myriad of gastrointestinal disorders, such as obscure gastrointestinal bleeding. This paper introduces a novel application of the semi-supervised learning approach for bleeding segmentation in VCE. A student U-Net equipped with a scSE attention block and a parallel teacher model are built, with their parameters synchronously updated during training. Our methodology is validated through experiments on the Kvasir-Capsule dataset, involving various GI bleeding conditions. The dataset's segmentation annotations have been manually provided by our team. The results confirm the efficacy of our SSL-based segmentation strategy in minimizing dependency on extensive annotations for model training, yet preserving the accuracy of bleeding recognition. Despite proposing an innovative computer-aided method for automated bleeding segmentation in VCE images – a potential alleviator of doctors' workloads – our approach is recommended for use as preliminary screening before manual diagnosis by a physician or for follow-up examination post-diagnosis.

\section*{Acknowledgements}
This work was supported by Hong Kong Research Grants Council (RGC) Research Impact Fund (RIF) R4020-22, Collaborative Research Fund (CRF C4026-21GF, CRF C4063-18G), General Research Fund (GRF 14203323),  NSFC/RGC Joint Research Scheme N\_CUHK420/22, GRS \#3110167; Shenzhen-Hong Kong-Macau Technology Research Programme (Type C) STIC Grant SGDX20210823103535014 (202108233000303); Guangdong Basic and Applied Basic Research Foundation (GBABF) \#2021B1515120035; Shun Hing Institute of Advanced Engineering (SHIAE Project BME-p1-21) at The Chinese University of Hong Kong (CUHK).

\bibliographystyle{cas-model2-names}

\bibliography{ref}

\clearpage

\normalMode







\end{document}